\documentclass[11pt]{article}

\usepackage[T1]{fontenc}
\usepackage[margin=1in]{geometry}
\usepackage{graphicx}
\usepackage{amsmath,amssymb}
\usepackage{booktabs}
\usepackage{multirow}
\usepackage{float}
\usepackage{algorithm}
\usepackage{algpseudocode}
\usepackage[numbers]{natbib}
\usepackage{hyperref}
\usepackage{xcolor}

\begin{document}

\title{RDGen: Demonstration Generation for High-Quality Robot Learning via Reinforcement Learning}
\author{%
Zijian Zhu\thanks{Equal contribution.}\quad
Menglin Zou\footnotemark[1]\quad
Zhuang Li\quad
Yaojie Tu\quad
Xinhai Sun\\
Synthoid.ai\\
\texttt{scholar@synthoid.ai}
}
\date{}

\maketitle

\begin{abstract}
Vision-Language-Action (VLA) models have emerged as a promising paradigm for general-purpose robot control. However, their performance remains fundamentally constrained by the availability of high-quality robot trajectory data. In current robot learning practice, such data are primarily collected through human teleoperation, which is labor-intensive, costly, and difficult to scale.
In this paper, we propose RDGen, a sim-to-real reinforcement learning framework for generating high-quality robot demonstrations. Rather than employing reinforcement learning solely as the final control policy, RDGen leverages trained RL policies as a structured trajectory generator. The system consists of a VLM-based task parser that identifies task-relevant objects, a Grounding DINO-based object localizer, and an RL policy transferred from simulation to the real robot. Successful rollouts are then harvested as clean, high-quality demonstrations for downstream VLA training, while the simulation stage further provides a scalable source of additional trajectories at little marginal cost.
Experiments on a pick-and-place task demonstrate that the transferred RL policy achieves a high task success rate. Compared with human teleoperation, RDGen produces significantly smoother trajectories and yields superior downstream VLA performance. These results indicate that RL-generated demonstrations can serve as more reliable and consistent supervisory signals for robot policy learning.

\end{abstract}

\noindent\textbf{Keywords:} Vision-Language-Action models, robot learning, reinforcement learning, sim-to-real transfer, robot data generation

\section{Introduction}

Recent advances in Vision-Language-Action (VLA) models have positioned them as a powerful approach to general-purpose robotic manipulation \cite{black2025pi0,bjorck2025gr00t,zitkovich2023rt2,kim2024openvla}. By conditioning policy outputs on visual observations, natural language instructions, and action histories, a VLA can map high-level semantic goals to low-level motor behavior in a unified framework. This multimodal formulation has made VLAs a compelling candidate for instruction-following manipulation, generalist robot policies, and scalable embodied learning.

Despite rapid progress in multimodal model design, the main bottleneck in VLA learning is increasingly a data bottleneck rather than a pure architecture. In contrast to text or image corpora, robot action data cannot be scraped at web scale. Each trajectory must be executed by a physical or simulated agent, grounded in a task, and paired with valid action supervision. As a result, the quantity, quality, and consistency of robot trajectories become first-order determinants of downstream policy performance~\cite{brohan2022rt}.

\begin{figure}[t]
\centering
\includegraphics[width=0.8\textwidth]{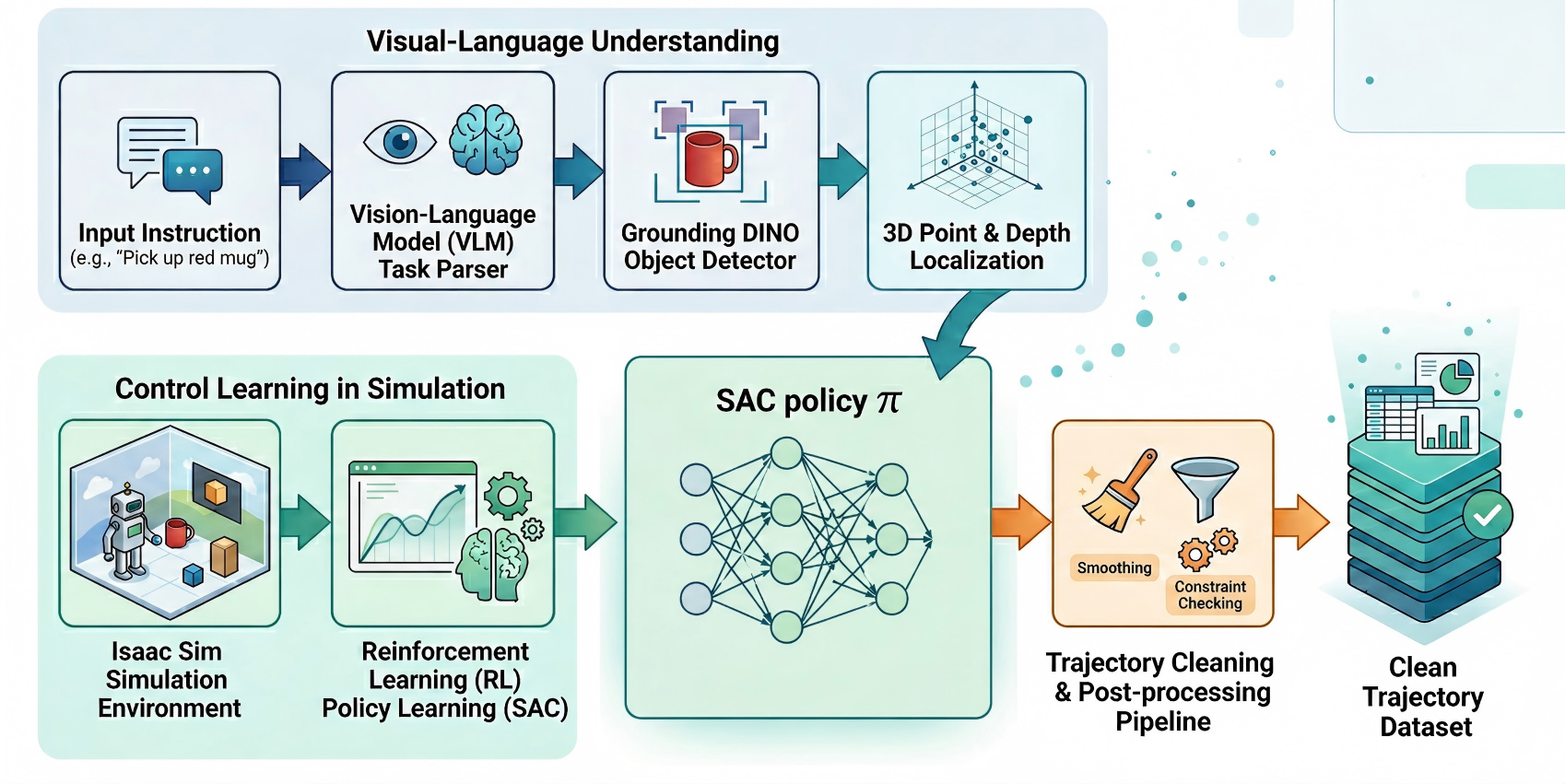}
\caption{
Overview of RDGen. Given a language instruction, a VLM parses the task and 
generates object queries, which are grounded into 3D positions via 
language-guided perception. Low-level control 
policies are trained in simulation using SAC and transferred to the real 
robot to execute tasks. Successful rollouts are collected and filtered to 
form a clean trajectory dataset, which serves as high-quality supervision 
for downstream VLA training.
}
\label{fig:pipeline}
\end{figure}

A prevalent approach to acquiring robot demonstrations is human teleoperation, which provides direct real-world task grounding and broad task coverage. Nevertheless, teleoperation remains inherently slow, costly, labor-intensive, and challenging to scale across diverse robotic platforms and environments \cite{mandlekar2018roboturk,khazatsky2024droid,zhao2023aloha,fu2024mobilealoha,mandlekar2021robomimic}. Egocentric human video offers a more scalable alternative for representation learning and task understanding \cite{grauman2022ego4d,kareer2024egomimic}, yet introduces an even larger embodiment gap. Critically, both sources share a fundamental limitation for action modeling: the resulting trajectories are inherently noisy, making it difficult to recover clean, robot-executable action labels for downstream policy training.


Our central insight is that reinforcement learning (RL), when integrated with an autonomous agent, can be transformed from a policy optimization method into an automated data-collection pipeline for generating high-quality robot trajectories. Once an RL policy successfully masters a task, it tends to produce trajectories that are goal-directed, physically executable, temporally consistent, and optimized for task completion under the specified reward structure. With appropriate reward shaping, such policies can further be encouraged to generate motions that are substantially smoother than human demonstrations. This motivates a reframing of RL's role in embodied learning: rather than serving as the deployed controller, RL can function as a data engine that supplies high-quality trajectories for downstream VLA supervision.


Motivated by these observations, we propose \textbf{RDGen}, a sim-to-real trajectory generation framework designed to supply clean, high-quality demonstrations for VLA training. RDGen integrates a Qwen3-VL-based task understanding agent for instruction parsing and target identification, a Grounding DINO-based depth perception module for 3D object localization, Soft Actor-Critic (SAC)-based policy learning for sim-to-real transfer, and an automated trajectory collection pipeline for downstream VLA supervision. The system is designed to transform high-level task descriptions into structured control problems, solve them via RL, and retain successful executions as clean, structured supervision signals.

In summary, this paper makes three key contributions. First, we introduce RDGen, a reinforcement learning-based demonstration generation paradigm that leverages RL as a data generator rather than solely as a deployed controller, producing high-quality robot trajectories for VLA training. Second, we develop an end-to-end automated sim-to-real trajectory generation pipeline that covers the full workflow from task understanding to real-world execution and data collection. Third, experiments on the target task show that RDGen collects high-quality, smoother trajectories that are easier for the model to fit than human teleoperation data, leading to better training performance.

\section{Background}
\subsection{Vision-Language-Action Models and Generalist Robot Policies}
Vision-Language-Action (VLA) models and generalist robot policies have recently emerged as a promising direction for scalable robot learning. RT-1 demonstrated that transformer-based robot policies can benefit from large-scale real-world robot data, while RT-2 showed that vision-language models can be co-trained with robot action data to transfer semantic knowledge into robotic control \cite{brohan2022rt,zitkovich2023rt2}. More recent systems, including GR00T N1, OpenVLA, and $\pi_0$, further suggest that broad task generalization depends not only on model capacity, but also on the scale, diversity, and quality of action-labeled robot trajectories \cite{bjorck2025gr00t,kim2024openvla,black2025pi0}. Most recently, GEN-0 and GEN-1 provide further evidence that embodied foundation models exhibit robotics-specific scaling laws, where increasing physical-interaction pretraining data and compute leads to predictable improvements in downstream robot performance \cite{generalist2025gen0,generalist2026gen1}. Our work is complementary to this line of research: instead of designing a new VLA architecture, we focus on improving the data side by generating high-quality robot demonstrations for VLA training.

\subsection{Robot Data Collection for Imitation and VLA Training}
Most large-scale robot learning datasets are built from human-collected robot demonstrations, typically obtained through teleoperation on real robot platforms. Datasets and systems such as robomimic, ALOHA, BridgeData V2, DROID, and Open X-Embodiment have significantly advanced robot imitation learning by providing real-world trajectories across diverse tasks, scenes, and embodiments \cite{zhao2023aloha,walke2023bridgedata,khazatsky2024droid,openxembodiment2023}. However, human-collected demonstrations are costly to acquire and often contain action noise caused by operator jitter, latency, hesitation, and corrective motions. Egocentric human video provides a more scalable source of embodied data for representation learning, pretraining, and task understanding \cite{grauman2022ego4d,kareer2024egomimic}; however, it introduces an even larger supervision gap for robot action learning because hand-tracking noise, camera ego-motion, occlusion, missing force and contact signals, and human-robot embodiment mismatch make it difficult to recover clean robot-executable action labels. In contrast, our goal is to generate trajectories that are not only grounded in task execution, but also smooth, successful, and directly usable as robot action supervision.

\subsection{Reinforcement Learning and Sim-to-Real Transfer}

Reinforcement learning (RL) formulates robot control as a sequential decision-making problem, commonly modeled as a Markov decision process (MDP) $\mathcal{M}=(\mathcal{S},\mathcal{A},p,r,\gamma)$, where $\mathcal{S}$ and $\mathcal{A}$ denote the state and action spaces, $p(s_{t+1}|s_t,a_t)$ defines the transition dynamics, $r(s_t,a_t)$ is the reward function, and $\gamma$ is the discount factor. An agent learns a policy $\pi_\theta(a|s)$ through interaction with the environment to maximize long-term task rewards.

RL has been widely applied to robotic continuous control and manipulation. Among common continuous-control algorithms, including DDPG, TD3, PPO, and Soft Actor-Critic (SAC) \cite{lillicrap2015continuous,fujimoto2018addressing,schulman2017proximal,haarnoja2018sac}, SAC is particularly attractive due to its off-policy learning, sample efficiency, and maximum-entropy objective. Specifically, it optimizes a policy by maximizing both expected task reward and policy entropy:
\begin{equation}
\pi^* = \arg\max_{\pi}
\mathbb{E}_{\tau\sim\pi}
\left[
\sum_{t=0}^{T}
\gamma^t
\left(
r(s_t,a_t) + \alpha \mathcal{H}(\pi(\cdot|s_t))
\right)
\right],
\label{eq:sac_return}
\end{equation}
where $\mathcal{H}(\pi(\cdot|s_t))$ denotes the entropy of the policy at state $s_t$, and $\alpha$ controls the trade-off between reward maximization and entropy regularization.

However, policies trained purely in simulation often suffer from the sim-to-real gap caused by discrepancies in dynamics, contacts, sensing, and visual appearance. To address this issue, sim-to-real methods such as domain randomization and adaptive simulation randomization expose policies to diverse simulated conditions during training, improving robustness when transferred to real robots \cite{tobin2017domain,chebotar2019simopt}.

Most prior sim-to-real RL methods aim to learn a controller that is directly deployed on the robot. In contrast, we use sim-to-real RL as an automated demonstration-generation mechanism. Once the learned policy can reliably complete a task, it produces successful, smooth, and robot-executable trajectories for VLA training. This reframes RL not as the final deployed policy, but as a scalable data-generation engine that complements human teleoperation and reduces the cost of collecting demonstrations.

\section{Demonstration Generation via Reinforcement Learning}

\subsection{High-quality Trajectories for VLA Training}

RDGen aims to generate high-quality robot demonstrations that can be used as supervision for VLA training. Each generated trajectory is represented as
\begin{equation}
\tau = \{(o_t, p_t, l, a_t)\}_{t=1}^{T},
\label{eq:traj}
\end{equation}
where $o_t$ denotes the visual observation at time step $t$, $p_t$ denotes the robot proprioceptive state, $l$ is the language instruction, and $a_t$ is the robot action.

Given a trajectory dataset $\mathcal{D} = \{\tau_i\}$ collected under our framework, a VLA policy learns to map multimodal context, including visual observations, natural language instructions, and robot proprioceptive states, to executable robot actions. Existing VLA models typically follow either autoregressive action prediction, where continuous actions are discretized into tokens~\cite{zitkovich2023rt2,kim2024openvla}, or continuous-action generation, where actions are modeled directly with diffusion- or flow-matching-based objectives~\cite{black2025pi0,shi2026saivla}. In both cases, the learned policy is strongly influenced by the quality of the action supervision provided by the trajectory data.

Noisy, inconsistent, or physically infeasible demonstrations introduce ambiguous training targets and can degrade the resulting policy. Therefore, RDGen retains only trajectories that satisfy both task-success and motion-quality criteria. Specifically, a trajectory is kept only if it completes the task and does not contain invalid states, unstable contacts, discontinuous commands, or physically non-executable motion. The resulting dataset $\mathcal{D}_{\mathrm{HQ}}$ provides clean, goal-directed, and robot-executable demonstrations for downstream VLA supervision.

\subsection{Task Understanding with VLM Agent}

We formulate language-driven long-horizon visual manipulation as a closed-loop agentic process driven by a Vision-Language Model. Given a user instruction $l$ and observation $o_t$, the agent repeatedly plans the next executable action, waits for the corresponding RL policy to execute it, verifies the outcome, and updates its task state. This design addresses the error accumulation and state drift commonly observed in long-horizon manipulation.

At time step $t$, the agent maintains a lightweight structured memory $m_t$ to
track the task progress. This memory records the original instruction, completed
actions, the remaining plan, and recent verification feedback. It is explicitly
stored and passed to each VLM call, allowing the agent to maintain consistent
task state across multiple planning and verification steps.

The action space is constrained by a closed atomic skill set, 
and each action is represented as
\begin{equation}
    a_t = \left(s_t, q_t^{\mathrm{obj}}, q_t^{\mathrm{tar}}\right),
    \label{eq:agent_action}
\end{equation}
where $s_t \in S$ denotes the skill type, $q_t^{\mathrm{obj}}$ denotes the
language description of the manipulated object, and $q_t^{\mathrm{tar}}$
denotes an optional language description of the target region. This representation maps VLM reasoning to executable robot skills and also provides object-level text prompts for the downstream localization module.

The framework contains two decoupled roles: a Planner and a Verifier.
Given the language instruction $l$, the current observation $o_t$, and the
structured memory $m_t$, the Planner generates the next high-level action in
the form of a skill, a manipulated-object query, and an optional target query. The Planner is prompted to focus on task-relevant visual evidence and to ignore the robot arm or gripper when identifying manipulable objects. This stage follows prior work on large-model-based task decomposition and embodied reasoning~\cite{huang2022language,liang2023code}.

After the RL policy executes the planned action $a_t$, the Verifier checks the
post-action observation $o_{t+1}$ and determines whether the intended effect has
been achieved. 
The task memory $m_t$ is updated according to the verification result. If the
Verifier confirms that the planned action has succeeded, the action is recorded
as completed and the agent proceeds to the next step. Otherwise, the memory is
kept unchanged, and the same step is retried or replanned. 


\subsection{Object Localization with Grounding DINO}

As described above, the VLM agent decomposes each task into a sequence of atomic actions and, for each action, identifies the relevant objects involved. These natural language descriptions are then passed to the perception module as queries, and the module uses them to localize the corresponding objects in the scene.

Given an RGB image $I_t$ and an agent-generated query
$q \in \{q_t^{\mathrm{obj}}, q_t^{\mathrm{tar}}\}$, Grounding DINO predicts a set of candidate bounding
boxes with confidence scores. We select the box with the highest confidence as
the object region:
\begin{equation}
    B_q = \arg\max_{B \in f_{\mathrm{DINO}}(I_t,q)} \mathrm{Conf}(B),
    \label{eq:grounding_dino_box}
\end{equation}
where $B_q$ denotes the selected bounding box for the queried object.

The selected 2D bounding box is then lifted to a 3D position using the aligned
depth map together with calibrated camera intrinsics and extrinsics. We denote
the resulting robot-frame 3D position of the queried object $q$ as
$\mathbf{p}_{r}^{q}$, where the subscript $r$ indicates the robot base frame.
Thus, for the manipulated object $q_t^{\mathrm{obj}}$ and the target object or region
$\tau_t$, the perception module estimates
\begin{equation}
    q \in \{q_t^{\mathrm{obj}}, q_t^{\mathrm{tar}}\}
    \quad \mapsto \quad
    \mathbf{p}_{r}^{q}.
    \label{eq:query_to_robot_position}
\end{equation}

Therefore, the symbolic action generated by the agent
is grounded as a metric robot command:
\begin{equation}
    \left(s_t, q_t^{\mathrm{obj}}, q_t^{\mathrm{tar}}\right)
    \quad \mapsto \quad
    \left(s_t, \mathbf{p}_{r}^{\mathrm{obj}}, \mathbf{p}_{r}^{\mathrm{tar}}\right).
    \label{eq:metric_action_grounding}
\end{equation}

\begin{figure}[t]
    \centering
    \includegraphics[width=0.5\linewidth]{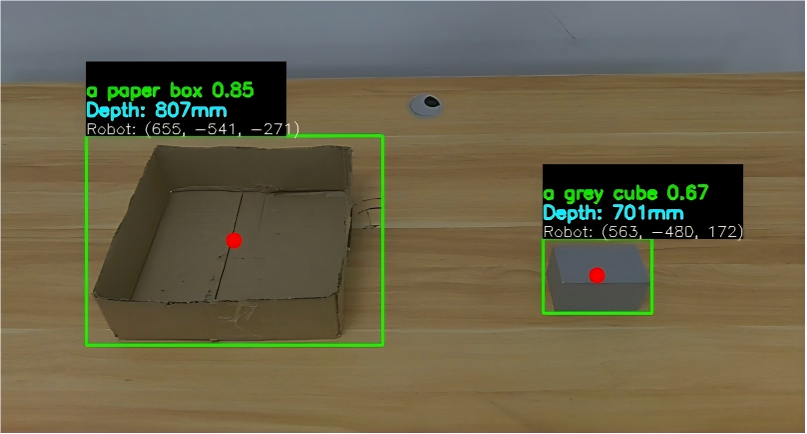}
    \caption{Example of language-guided object localization in a real scene. 
Grounding DINO detects the target object ``paper box'' and the manipulated
object ``gray cube'' from agent-generated language queries.}
    \label{fig:sim2real_grounding}
\end{figure}

This module serves as a lightweight sim-to-real grounding interface. In
simulation, object and target coordinates are directly available from the
environment state, whereas in the real world they are estimated from
language-guided detection and RGB-D observations. As shown in
Fig.~\ref{fig:sim2real_grounding}, this allows the same object-level skill
policy to use real-world target coordinates during physical execution.

Such a design enables strong generalization, as the system can localize arbitrary objects directly from language queries without requiring object-specific geometric models, such as CAD models or object-level point clouds \cite{wang2019densefusion,xiang2017posecnn}.

\subsection{Policy Learning}

Robotic manipulation with dexterous hands requires coordinated arm control, stable grasping, and smooth motion generation. We formulate each low-level manipulation skill as a reinforcement learning problem and train the policy in Isaac Sim, where the robot embodiment, workspace, and task objects are modeled to provide a controllable environment for policy optimization.

\textbf{Observation Space.}
At each time step $t$, the policy state is represented by the target position, the end-effector position, and the end-effector orientation. To address the sign ambiguity of quaternion representations, both the original quaternion and its negated equivalent are included in the state representation. The state vector is defined as:
\begin{equation}
\mathbf{s}_t =
\left[
\mathbf{p}_{\mathrm{target},t},
\mathbf{p}_{\mathrm{ee},t},
\mathbf{q}_{\mathrm{ee},t}^{+},
\mathbf{q}_{\mathrm{ee},t}^{-}
\right],
\end{equation}
where $\mathbf{p}_{\mathrm{target},t} \in \mathbb{R}^{3}$ denotes the target position, $\mathbf{p}_{\mathrm{ee},t} \in \mathbb{R}^{3}$ denotes the end-effector position, $\mathbf{q}_{\mathrm{ee},t}^{+} = [w_t, x_t, y_t, z_t]^\top \in \mathbb{R}^{4}$ denotes the measured end-effector orientation quaternion, and $\mathbf{q}_{\mathrm{ee},t}^{-} = -\mathbf{q}_{\mathrm{ee},t}^{+}$ denotes its sign-equivalent representation. Therefore, the complete state vector has dimension $14$.

\textbf{Action Space.}
The action is defined in the end-effector pose space. The policy outputs the desired end-effector motion command, including Cartesian position control and orientation control represented in Euler angles. The commanded end-effector pose is then converted to joint-space motion through inverse kinematics and executed by the robot controller.

\begin{equation}
\mathbf{a}_t =
\left[
\Delta x_{\mathrm{ee}},
\Delta y_{\mathrm{ee}},
\Delta z_{\mathrm{ee}},
\Delta \phi_{\mathrm{ee}},
\Delta \theta_{\mathrm{ee}},
\Delta \psi_{\mathrm{ee}}
\right]^\top
\in \mathbb{R}^{6},
\end{equation}

\textbf{Training.}
A unified pick-and-place skill is implemented as a low-level reinforcement learning policy trained in simulation. During training, the environment provides structured task information, including the target object pose, placement target pose, robot state, and end-effector pose. The policy takes this observation as input and predicts an end-effector pose command, which is converted into executable joint targets through inverse kinematics and applied to the simulated robot.

We optimize the policy using Soft Actor-Critic (SAC). In this framework, SAC learns the low-level control behavior required for pick-and-place execution, while the task-specific reward and termination conditions define the desired manipulation outcome, such as reaching the object, grasping it, lifting it, and placing it at the target location.

The training process follows a closed-loop interaction cycle. At each step, the policy observes the current robot and task state, predicts an end-effector command, and executes it through inverse kinematics. The resulting transition is then used to update the policy.

To improve robustness, we randomize object and target positions within the feasible workspace and inject noise into robot observations during training. After training, successful policy rollouts are collected as clean, goal-directed demonstrations for downstream VLA training.


\textbf{Reward.}
The reward function balances task completion and trajectory quality. It combines task-specific rewards with shaping terms that encourage progress and target alignment, while penalizing invalid or inefficient behaviors.
Thus,
the complete reward at time step $t$ is expressed as
\begin{equation}
r_t =
r_{\mathrm{task}}
+
r_{\mathrm{shape}}
-
r_{\mathrm{penalty}} ,
\end{equation}

where
\begin{equation}
r_{\mathrm{penalty}}
=
\lambda_{\mathrm{still}} \mathbb{I}_{\mathrm{still}}
+
\lambda_{\mathrm{step}}
+
\lambda_{\mathrm{IK}} \mathbb{I}_{\mathrm{IK\_fail}}
+
\lambda_{\mathrm{collision}} \mathbb{I}_{\mathrm{collision}} ,
\end{equation}
where $\mathbb{I}_{\mathrm{still}}$, $\mathbb{I}_{\mathrm{IK\_fail}}$, and
$\mathbb{I}_{\mathrm{collision}}$ are binary indicators for stationary motion,
inverse-kinematics failure, and collision, respectively. The coefficients
$\lambda_{\mathrm{still}}$, $\lambda_{\mathrm{IK}}$, and
$\lambda_{\mathrm{collision}}$ control the strength of each penalty, while
$\lambda_{\mathrm{step}}$ is a constant per-step penalty that discourages
unnecessarily long episodes.

The shaping reward is defined as:
\begin{equation}
r_{\mathrm{shape}}
=
r_{\mathrm{progress}}
+
r_{\mathrm{center}}
+
r_{\mathrm{path}} ,
\end{equation}

The progress reward encourages the end-effector to reduce its distance to the target between two consecutive time steps:
\begin{equation}
r_{\mathrm{progress}} =
\lambda_{\mathrm{progress}}
\left(
d_{t-1} - d_t
\right),
\end{equation}
where $d_{t-1}$ and $d_t$ denote the end-effector-to-target distance at the previous and current time steps, respectively.

The distance-centering reward encourages coordinate-wise movement toward the target. Since different motion directions may contribute differently to task success, separate weights are assigned to the $x$, $y$, and $z$ directions:
\begin{equation}
\label{eq:center}
r_{\mathrm{center}} =
\boldsymbol{\lambda}_{xyz}^{\top}
\left(
\left|\mathbf{p}_{t-1} - \mathbf{p}_{\mathrm{target}}\right|
-
\left|\mathbf{p}_{t} - \mathbf{p}_{\mathrm{target}}\right|
\right),
\end{equation}
where $\mathbf{p}_{t-1}$ and $\mathbf{p}_{t}$ denote the previous and current end-effector positions, $\mathbf{p}_{\mathrm{target}}$ denotes the target position, and $\boldsymbol{\lambda}_{xyz} = [\lambda_x,\lambda_y,\lambda_z]^\top$ contains the coordinate-wise reward weights. This term becomes positive when the end-effector reduces its coordinate-wise distance to the target.

We also use a path-deviation penalty to improve trajectory quality, which penalizes the distance between the current end-effector position and the straight-line segment connecting the episode start position and the target position. 
The path-deviation reward is defined as
\begin{equation}
r_{\mathrm{path}} =
-\lambda_{\mathrm{path}}
\left\|
\mathbf{p}_{\mathrm{ee}} -
\mathbf{p}_{\mathrm{proj}}
\right\|_2 ,
\end{equation}
where $\lambda_{\mathrm{path}}$ controls the strength of the penalty and
$\mathbf{p}_{\mathrm{proj}}$ denotes the closest point on the desired
start-to-target path. This term penalizes the distance between the current
end-effector position $\mathbf{p}_{\mathrm{ee}}$ and the projected point
$\mathbf{p}_{\mathrm{proj}}$, encouraging the end-effector to stay close to a
direct trajectory.

The projected point $\mathbf{p}_{\mathrm{proj}}$ is computed as
\begin{equation}
\mathbf{p}_{\mathrm{proj}} =
\mathbf{p}_{\mathrm{start}}
+
\mathrm{clip}\left(
\frac{
(\mathbf{p}_{\mathrm{ee}} - \mathbf{p}_{\mathrm{start}})^\top \mathbf{s}
}{
\|\mathbf{s}\|_2^2
},
0, 1
\right)\mathbf{s}.
\end{equation}
where $\mathbf{p}_{\mathrm{start}}$ is the episode start position and
$\mathbf{s}$ is the direction vector of the desired path.



\textbf{Reset Strategy.}
We use early termination and reset to make training more sample-efficient. Rather
than allowing an episode to continue after the robot reaches an invalid or
unrecoverable state, the environment immediately terminates the episode and
starts a new rollout. This focuses training on meaningful interaction data and
reduces the amount of experience collected from states that are unlikely to
lead to task success.

An episode is terminated when any of the following conditions is satisfied:
\begin{equation}
\mathbb{I}_{\mathrm{done}}
=
\mathbb{I}_{\mathrm{success}}
\vee
\mathbb{I}_{\mathrm{table}}
\vee
\mathbb{I}_{\mathrm{target}}
\vee
\mathbb{I}_{\mathrm{IK}}
\vee
\mathbb{I}_{t \geq T_{\max}} .
\end{equation}
Here, $\mathbb{I}_{\mathrm{success}}$ denotes successful task completion,
$\mathbb{I}_{\mathrm{table}}$ indicates unsafe contact with the table,
$\mathbb{I}_{\mathrm{target}}$ indicates undesired movement of the target
object, $\mathbb{I}_{\mathrm{IK}}$ indicates repeated inverse-kinematics
failure, and $T_{\max}$ is the maximum episode horizon.

After termination, the simulator resets the robot, object, and task state. The
robot is returned to a safe initial configuration, and the target object is
resampled within the feasible workspace. This reset procedure exposes the policy
to diverse task configurations while preventing training from being dominated by
invalid contacts, failed IK commands, or excessively long unsuccessful
episodes.


\subsection{Sim-to-Real Adaptation}

To transfer the simulation-trained RL policy to the real robot, the training observation structure was improved to better match the noise and ambiguity present in real-world sensing. During training, random noise was added to the end-effector pose, including $5\,\mathrm{mm}$ perturbations in the $x$, $y$, and $z$ position directions, and $0.1$ perturbations to the quaternion orientation components $(w, x, y, z)$, followed by quaternion normalization.

A key challenge in real-robot deployment is that quaternion representations are sign-ambiguous: $q$ and $-q$ represent the same physical orientation. However, in practice, sensor readings and forward kinematics calculations may randomly output either $q$ or $-q$, which can make the observation unstable and confusing for policy learning. To address this issue, a double-quaternion representation was introduced into the observation space by including both the original quaternion $q=(w,x,y,z)$ and its negated form $-q=(-w,-x,-y,-z)$. This makes the policy robust to quaternion sign flips in the real system.

Therefore, the final observation is a $14$-dimensional vector consisting of the target position, end-effector position, quaternion orientation, and negative quaternion orientation:
\begin{equation}
\mathbf{o}_t =
\left[
\mathbf{p}_{\mathrm{target}},
\mathbf{p}_{\mathrm{ee}},
q_{\mathrm{ee}},
-q_{\mathrm{ee}}
\right].
\end{equation}

Specifically, this vector contains the target position, the end-effector position, the orientation quaternion, and the negated orientation quaternion. Depending on whether the original quaternion has a positive or negative $w$ value, the corresponding quaternion representation is placed in its appropriate observation position. This ordering strategy reduces the ambiguity caused by the fact that a quaternion $q$ and its negated form $-q$ represent the same physical orientation. Without a consistent placement rule, the same robot pose could appear in the observation either as $[q,-q]$ or $[-q,q]$, leading to discontinuous state representations and causing the policy network to interpret identical poses as different states. By assigning the quaternion order according to the sign of $w$, the observation becomes more temporally consistent, which improves training stability and enables the simulation-trained RL policy to be deployed more reliably on the real robot.

\subsection{Scalable Trajectory Collection in Simulation}

Beyond real-robot deployment, RDGen also leverages simulation as a scalable source of trajectory data. By varying environmental factors such as lighting conditions, object poses, camera viewpoints, and scene configurations, RDGen can generate diverse successful trajectories at low cost. These simulated trajectories can be used as pretraining data for downstream VLA models, improving data coverage before real-robot demonstration collection.

\section{Experiments}
We evaluate RDGen in terms of trajectory quality and motion smoothness. 
Trajectory quality is evaluated by the downstream VLA success rate, which reflects whether the generated trajectories provide useful training data for policy learning. 

Motion smoothness measures unnecessary oscillations and abrupt motion changes. According to the minimum-jerk trajectory principle~\cite{sharkawy2021minimum}, smooth motion can be characterized by minimizing the variation of acceleration over time. Specifically, jerk is defined as the time derivative of acceleration:
\begin{equation}
    \mathbf{j}(t) = \frac{d\mathbf{a}(t)}{dt}
    = \frac{d^3\mathbf{x}(t)}{dt^3},
    \label{eq:jerk-definition}
\end{equation}
and smooth trajectories can be obtained by minimizing the squared jerk:
\begin{equation}
    J = \int_{0}^{t_f} \left\| \mathbf{j}(t) \right\|^2 dt.
    \label{eq:jerk-objective}
\end{equation}

In our experiments, we compute the average jerk magnitude for each recorded trajectory and report the mean value across all trajectories as the smoothness metric. A lower value indicates smoother motion with fewer abrupt changes and less oscillation during execution.



\subsection{Experimental Setup}

The experimental platform is described in Appendix~\ref{app:hardware-software-settings}. To ensure consistency between simulation and real-world deployment, the simulation environment is constructed in Isaac Sim using the same configuration.

Both the simulation and real-world environments share a unified coordinate system. In this coordinate frame, the positive y-axis points upward, the positive z-axis points to the right, and the positive x-axis points forward. This alignment reduces the coordinate transformation gap between simulation and the real robot, enabling more direct policy transfer.

We collect 20 trajectories to evaluate trajectory smoothness using the average jerk metric. For downstream VLA evaluation, we train a $\pi_0$ policy \cite{black2025pi0} using the collected trajectories and report the success rate over 10 trials with randomly initialized cube and box positions.

\subsection{Results}

\begin{figure}[t]
    \centering
    \includegraphics[width=0.9\linewidth]{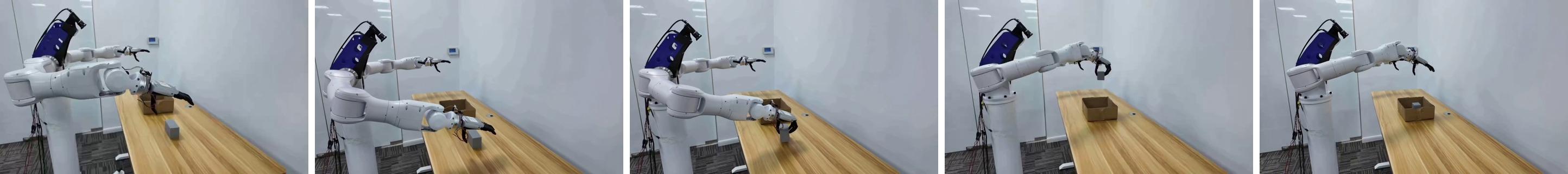}
    \caption{Task execution of grasping a gray cube and placing it into a paper box.}
    \label{fig:task-success}
\end{figure}

The RL policy achieves a 100\% success rate on the task of grasping a gray cube and placing it into a paper box, as shown in Fig.~\ref{fig:task-success}, where all 20 collected rollouts succeeded on the first attempt without any manual intervention.

We further compare the trajectory smoothness and downstream VLA performance between teleoperated data and RDGen-generated data. The results are summarized in Table~\ref{tab:jerk-vla}.

\begin{table}[h]
    \centering
    \caption{Comparison of trajectory smoothness and downstream VLA training performance across different tasks. Mean jerk is averaged over 20 trajectories per method. VLA performance is evaluated over 10 trials with different object positions.}
    \label{tab:jerk-vla}
    \begin{tabular}{llcc}
        \hline
        Task & Method & Mean Jerk (m/s$^3$) $\downarrow$ & VLA Success Rate $\uparrow$ \\
        \hline
        Cube & Teleoperation & $2.68 \pm 0.41$ & 60\% \\
        Cube & RDGen & $0.47 \pm 0.02$ & 80\% \\
        Cola & Teleoperation & $5.59 \pm 2.44$ & 70\% \\
        Cola & RDGen & $0.57 \pm 0.06$ & 100\% \\
        \hline
    \end{tabular}
\end{table}

Table~\ref{tab:jerk-vla} compares the two demonstration sources along two complementary axes. 
In terms of trajectory smoothness, RDGen consistently produces lower Cartesian jerk at the end-effector, 
indicating that the generated motions are smoother and less noisy than those collected via teleoperation. 
This improvement also translates to downstream policy learning, where the VLA model trained on RDGen data 
achieves a higher task success rate. 
Taken together, these results suggest that RDGen produces higher-quality demonstrations that improve both 
motion smoothness and downstream learnability.

\section{Conclusion}
In this paper, we presented RDGen, a reinforcement learning-based framework for generating high-quality robot demonstrations for downstream VLA training. RDGen uses sim-to-real RL policies as trajectory generators rather than only as deployed controllers, enabling the collection of successful, smooth, and physically executable demonstrations. By combining VLM-based task understanding, language-guided object localization, SAC-based skill learning, and real-world execution, the system provides an automated pipeline from natural language instructions to robot trajectory data.

Despite these promising results, RDGen still has several limitations. First, the current framework is primarily designed for relatively coarse manipulation tasks such as grasping and placement, and may not readily extend to fine-grained skills — such as garment folding or screw tightening — that demand millimeter-level precision and require carefully engineered reward functions for effective reinforcement learning training. Second, the training pipeline is not yet fully task-general: some design choices, such as using a spring mechanism to randomize object positions in grasping tasks, are task-specific and may not directly transfer to other manipulation skills. Future work will improve support for fine-grained manipulation and develop more general automated training mechanisms for diverse robot tasks.

\renewcommand{\bibname}{References}
\bibliographystyle{splncs04}
\bibliography{references}

@inproceedings{black2025pi0,
  title     = {{$\pi_0$}: A Vision-Language-Action Flow Model for General Robot Control},
  author    = {Black, Kevin and Brown, Noah and Driess, Danny and Esmail, Adnan and Equi, Michael and Finn, Chelsea and Fusai, Niccolo and Groom, Lachy and Hausman, Karol and Ichter, Brian and Jakubczak, Szymon and Jones, Tim and Ke, Liyiming and Levine, Sergey and Li-Bell, Adrian and others},
  booktitle = {Robotics: Science and Systems},
  year      = {2025}
}

@inproceedings{zitkovich2023rt2,
  title={Rt-2: Vision-language-action models transfer web knowledge to robotic control},
  author={Zitkovich, Brianna and Yu, Tianhe and Xu, Sichun and Xu, Peng and Xiao, Ted and Xia, Fei and Wu, Jialin and Wohlhart, Paul and Welker, Stefan and Wahid, Ayzaan and others},
  booktitle={Conference on Robot Learning},
  pages={2165--2183},
  year={2023},
  organization={PMLR}
}

@article{brohan2022rt,
  title={Rt-1: Robotics transformer for real-world control at scale},
  author={Brohan, Anthony and Brown, Noah and Carbajal, Justice and Chebotar, Yevgen and Dabis, Joseph and Finn, Chelsea and Gopalakrishnan, Keerthana and Hausman, Karol and Herzog, Alex and Hsu, Jasmine and others},
  journal={arXiv preprint arXiv:2212.06817},
  year={2022}
}

@article{bjorck2025gr00t,
  title={Gr00t n1: An open foundation model for generalist humanoid robots},
  author={Bjorck, Johan and Casta{\~n}eda, Fernando and Cherniadev, Nikita and Da, Xingye and Ding, Runyu and Fan, Linxi and Fang, Yu and Fox, Dieter and Hu, Fengyuan and Huang, Spencer and others},
  journal={arXiv preprint arXiv:2503.14734},
  year={2025}
}

@article{kim2024openvla,
  title={Openvla: An open-source vision-language-action model},
  author={Kim, Moo Jin and Pertsch, Karl and Karamcheti, Siddharth and Xiao, Ted and Balakrishna, Ashwin and Nair, Suraj and Rafailov, Rafael and Foster, Ethan and Lam, Grace and Sanketi, Pannag and others},
  journal={arXiv preprint arXiv:2406.09246},
  year={2024}
}

@inproceedings{walke2023bridgedata,
  title={Bridgedata v2: A dataset for robot learning at scale},
  author={Walke, Homer Rich and Black, Kevin and Zhao, Tony Z and Vuong, Quan and Zheng, Chongyi and Hansen-Estruch, Philippe and He, Andre Wang and Myers, Vivek and Kim, Moo Jin and Du, Max and others},
  booktitle={Conference on Robot Learning},
  pages={1723--1736},
  year={2023},
  organization={PMLR}
}

@inproceedings{zhao2023aloha,
  title     = {Learning Fine-Grained Bimanual Manipulation with Low-Cost Hardware},
  author    = {Zhao, Tony Z. and Kumar, Vikash and Levine, Sergey and Finn, Chelsea},
  booktitle = {Robotics: Science and Systems},
  year      = {2023}
}

@inproceedings{fu2024mobilealoha,
  title     = {Mobile {ALOHA}: Learning Bimanual Mobile Manipulation with Low-Cost Whole-Body Teleoperation},
  author    = {Fu, Zipeng and Zhao, Tony Z. and Finn, Chelsea},
  booktitle = {Proceedings of the Conference on Robot Learning},
  year      = {2024}
}

@article{khazatsky2024droid,
  title={Droid: A large-scale in-the-wild robot manipulation dataset},
  author={Khazatsky, Alexander and Pertsch, Karl and Nair, Suraj and Balakrishna, Ashwin and Dasari, Sudeep and Karamcheti, Siddharth and Nasiriany, Soroush and Srirama, Mohan Kumar and Chen, Lawrence Yunliang and Ellis, Kirsty and others},
  journal={arXiv preprint arXiv:2403.12945},
  year={2024}
}

@inproceedings{mandlekar2021robomimic,
  title     = {{robomimic}: A Benchmark for Robot Learning from Demonstration},
  author    = {Mandlekar, Ajay and Xu, Danfei and Wong, Josiah and Nasiriany, Soroush and Wang, Chen and Kulkarni, Rohun and Fei-Fei, Li and Savarese, Silvio and Zhu, Yuke and Mart{\'i}n-Mart{\'i}n, Roberto},
  booktitle = {Conference on Robot Learning},
  year      = {2021}
}

@inproceedings{mandlekar2018roboturk,
  title     = {{RoboTurk}: A Crowdsourcing Platform for Robotic Skill Learning through Imitation},
  author    = {Mandlekar, Ajay and Zhu, Yuke and Garg, Animesh and Fei-Fei, Li and Savarese, Silvio},
  booktitle = {Conference on Robot Learning},
  year      = {2018}
}

@inproceedings{grauman2022ego4d,
  title={Ego4d: Around the world in 3,000 hours of egocentric video},
  author={Grauman, Kristen and Westbury, Andrew and Byrne, Eugene and Chavis, Zachary and Furnari, Antonino and Girdhar, Rohit and Hamburger, Jackson and Jiang, Hao and Liu, Miao and Liu, Xingyu and others},
  booktitle={Proceedings of the IEEE/CVF conference on computer vision and pattern recognition},
  pages={18995--19012},
  year={2022}
}

@article{kareer2024egomimic,
  title   = {{EgoMimic}: Scaling Imitation Learning via Egocentric Video},
  author  = {Kareer, Simar and Patel, Dhruv and Punamiya, Ryan and Mathur, Pranay and Cheng, Shuo and Wang, Chen and Hoffman, Judy and Xu, Danfei},
  journal = {arXiv preprint arXiv:2410.24221},
  year    = {2024}
}

@article{openxembodiment2023,
  title   = {Open {X}-Embodiment: Robotic Learning Datasets and {RT-X} Models},
  author  = {{Open X-Embodiment Collaboration} and Abou-Chakra, Jad and others},
  journal = {arXiv preprint arXiv:2310.08864},
  year    = {2023}
}

@inproceedings{huang2022language,
  title={Language models as zero-shot planners: Extracting actionable knowledge for embodied agents},
  author={Huang, Wenlong and Abbeel, Pieter and Pathak, Deepak and Mordatch, Igor},
  booktitle={International conference on machine learning},
  pages={9118--9147},
  year={2022},
  organization={PMLR}
}

@inproceedings{liang2023code,
  title={Code as policies: Language model programs for embodied control},
  author={Liang, Jacky and Huang, Wenlong and Xia, Fei and Xu, Peng and Hausman, Karol and Ichter, Brian and Florence, Pete and Zeng, Andy},
  booktitle={2023 IEEE International conference on robotics and automation (ICRA)},
  pages={9493--9500},
  year={2023},
  organization={IEEE}
}

@article{generalist2025gen0,
  author = {Generalist AI Team},
  title = {GEN-0: Embodied Foundation Models That Scale with Physical Interaction},
  journal = {Generalist AI Blog},
  year = {2025},
  note = {https://generalistai.com/blog/nov-04-2025-GEN-0}
}

@article{generalist2026gen1,
  author = {Generalist AI Team},
  title = {GEN-1: Scaling Embodied Foundation Models to Mastery},
  journal = {Generalist AI Blog},
  year = {2026},
  note = {https://generalistai.com/blog/apr-02-2026-GEN-1}
}

@inproceedings{haarnoja2018sac,
  title={Soft Actor-Critic: Off-Policy Maximum Entropy Deep Reinforcement Learning with a Stochastic Actor},
  author={Haarnoja, Tuomas and Zhou, Aurick and Abbeel, Pieter and Levine, Sergey},
  booktitle={Proceedings of the International Conference on Machine Learning},
  pages={1861--1870},
  year={2018},
  organization={PMLR}
}

@article{lillicrap2015continuous,
  title={Continuous Control with Deep Reinforcement Learning},
  author={Lillicrap, Timothy P. and Hunt, Jonathan J. and Pritzel, Alexander and Heess, Nicolas and Erez, Tom and Tassa, Yuval and Silver, David and Wierstra, Daan},
  journal={arXiv preprint arXiv:1509.02971},
  year={2015}
}

@inproceedings{fujimoto2018addressing,
  title={Addressing Function Approximation Error in Actor-Critic Methods},
  author={Fujimoto, Scott and van Hoof, Herke and Meger, David},
  booktitle={Proceedings of the International Conference on Machine Learning},
  pages={1587--1596},
  year={2018},
  organization={PMLR}
}

@article{schulman2017proximal,
  title={Proximal Policy Optimization Algorithms},
  author={Schulman, John and Wolski, Filip and Dhariwal, Prafulla and Radford, Alec and Klimov, Oleg},
  journal={arXiv preprint arXiv:1707.06347},
  year={2017}
}

@inproceedings{tobin2017domain,
  title={Domain Randomization for Transferring Deep Neural Networks from Simulation to the Real World},
  author={Tobin, Josh and Fong, Rachel and Ray, Alex and Schneider, Jonas and Zaremba, Wojciech and Abbeel, Pieter},
  booktitle={Proceedings of the IEEE/RSJ International Conference on Intelligent Robots and Systems},
  pages={23--30},
  year={2017}
}

@inproceedings{chebotar2019simopt,
  title={Closing the Sim-to-Real Loop: Adapting Simulation Randomization with Real World Experience},
  author={Chebotar, Yevgen and Handa, Ankur and Makoviychuk, Viktor and Macklin, Miles and Issac, Jan and Ratliff, Nathan and Fox, Dieter},
  booktitle={Proceedings of the IEEE International Conference on Robotics and Automation},
  pages={8973--8979},
  year={2019}
}

@article{shi2026saivla,
  title={SaiVLA-0: Cerebrum--Pons--Cerebellum Tripartite Architecture for Compute-Aware Vision-Language-Action},
  author={Shi, Xiang and Huang, Wenlong and Zou, Menglin and Sun, Xinhai},
  journal={arXiv preprint arXiv:2603.08124},
  year={2026}
}

@article{xiang2017posecnn,
  title={Posecnn: A convolutional neural network for 6d object pose estimation in cluttered scenes},
  author={Xiang, Yu and Schmidt, Tanner and Narayanan, Venkatraman and Fox, Dieter},
  journal={arXiv preprint arXiv:1711.00199},
  year={2017}
}

@inproceedings{wang2019densefusion,
  title={Densefusion: 6d object pose estimation by iterative dense fusion},
  author={Wang, Chen and Xu, Danfei and Zhu, Yuke and Mart{\'\i}n-Mart{\'\i}n, Roberto and Lu, Cewu and Fei-Fei, Li and Savarese, Silvio},
  booktitle={Proceedings of the IEEE/CVF conference on computer vision and pattern recognition},
  pages={3343--3352},
  year={2019}
}

@article{sharkawy2021minimum,
  title={Minimum jerk trajectory generation for straight and curved movements: Mathematical analysis},
  author={Sharkawy, Abdel-Nasser},
  journal={arXiv preprint arXiv:2102.07459},
  year={2021}
}

\appendix

\section{Experimental Settings and Additional Results}
\label{app:experimental-settings-additional-results}

This appendix provides the detailed experimental settings used in our experiments, including the training hyperparameters, hardware and software configuration, and additional qualitative results.

\subsection{Experimental Setup}
\label{app:training-hyperparameters}
\label{app:hardware-software-settings}

\begin{table}[H]
    \centering
    \caption{Training hyperparameters used for RL training.}
    \label{tab:training-hyperparameters}
    \begin{tabular}{l c}
        \hline
        Parameter & Value \\
        \hline
        Learning rate & $1\times10^{-4}$ \\
        Batch size & 256 \\
        Gradient steps & 16 \\
        Temperature & 0.2 \\
        \hline
    \end{tabular}
\end{table}

For the downstream VLA training, we followed the same hyperparameter configuration as the adopted $\pi_0$ without modification.

\vspace{-0.5em}

\begin{table}[H]
    \centering
    \caption{Hardware and software environment.}
    \label{tab:hardware-software-settings}
    \begin{tabular}{l c}
        \hline
        Item & Configuration \\
        \hline
        Robot Arm & Marvin M6CCS \\
        Dexterous Hand & Ruiyan RY-H2 \\
        Camera & Gemini 435Le \\
        GPU & NVIDIA RTX PRO 6000 \\
        \hline
    \end{tabular}
\end{table}

\vspace{-0.5em}

\subsection{Additional Results}
\label{app:additional-results}

\begin{figure}[H]
    \centering
    \includegraphics[width=0.75\linewidth]{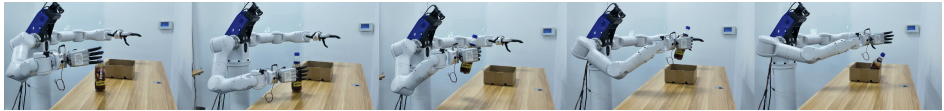}
    \caption{Qualitative result of grasping a bottle and placing it into a paper box.}
    \label{fig:additional-result}
\end{figure}

\end{document}